\documentclass{article}

\usepackage{microtype}
\usepackage{graphicx}
\usepackage{xcolor}

\usepackage{subfigure}
\usepackage{booktabs} % for professional tables
\usepackage{url}
\usepackage{multirow}

\usepackage{amsmath,amsfonts,bm}
\usepackage{microtype}
\usepackage{graphicx}
\usepackage{booktabs} % for professional tables
\usepackage{times}
\usepackage{epsfig}
\usepackage{graphicx}
\usepackage{amssymb}
\usepackage{enumitem}

\def\eqref#1{equation~\ref{#1}}
\def\1{\bm{1}}

\DeclareMathAlphabet{\mathsfit}{\encodingdefault}{\sfdefault}{m}{sl}
\SetMathAlphabet{\mathsfit}{bold}{\encodingdefault}{\sfdefault}{bx}{n}

\newcommand{\norm}[1]{\left\lVert#1\right\rVert}

\usepackage{hyperref}

\usepackage[accepted]{icml2021}

\icmltitlerunning{VideoGPT}

\begin{document}

\twocolumn[
\icmltitle{VideoGPT: Video Generation using VQ-VAE and Transformers}

\icmlsetsymbol{equal}{*}

\begin{icmlauthorlist}
\icmlauthor{Wilson Yan}{equal,be}
\icmlauthor{Yunzhi Zhang}{equal,be}
\icmlauthor{Pieter Abbeel}{be}
\icmlauthor{Aravind Srinivas}{be}
\end{icmlauthorlist}

\icmlaffiliation{be}{University of California, Berkeley}

\icmlcorrespondingauthor{Wilson Yan, Aravind Srinivas}{wilson1.yan@berkeley.edu, aravind\_srinivas@berkeley.edu}

\icmlkeywords{Machine Learning, ICML}

\vskip 0.3in
]

\printAffiliationsAndNotice{\icmlEqualContribution} % otherwise use the standard text.

\begin{figure*}[!ht]
    \centering
    \includegraphics[width=\textwidth]{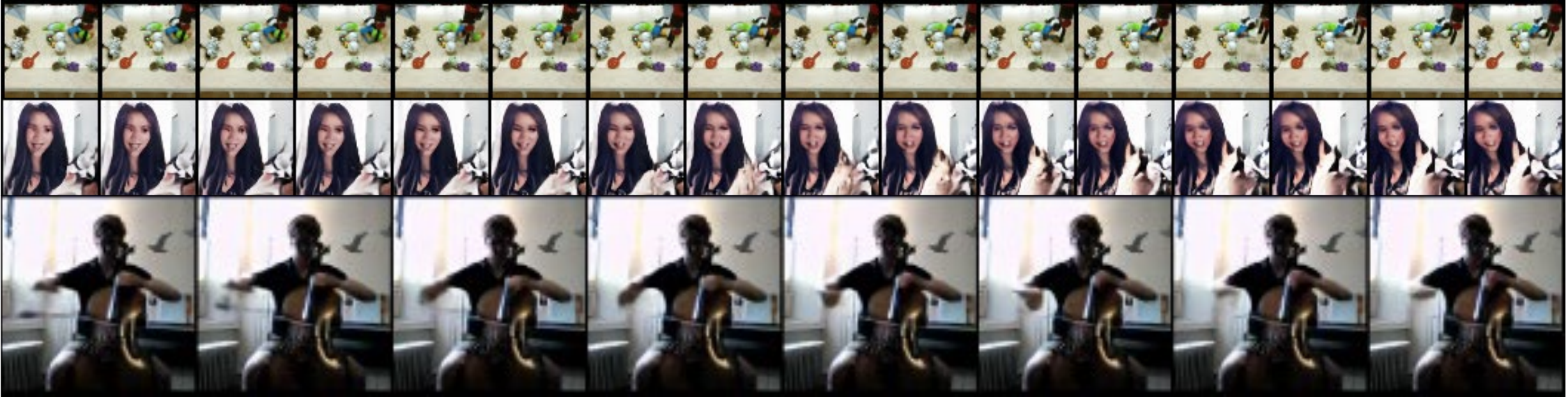}
    \caption{$64\times 64$ and $128\times 128$ video samples generated by VideoGPT}
    \label{fig:intro}
\end{figure*}

\begin{abstract}
We present VideoGPT: a conceptually simple architecture for scaling likelihood based generative modeling to natural videos. VideoGPT uses VQ-VAE that learns downsampled discrete latent representations of a raw video by employing 3D convolutions and axial self-attention. A simple GPT-like architecture is then used to autoregressively model the discrete latents using spatio-temporal position encodings. Despite the simplicity in formulation and ease of training, our architecture is able to generate samples competitive with state-of-the-art GAN models for video generation on the BAIR Robot dataset, and generate high fidelity natural videos from UCF-101 and Tumbler GIF Dataset (TGIF). We hope our proposed architecture serves as a reproducible reference for a minimalistic implementation of transformer based video generation models. Samples and code are available at \url{https://wilson1yan.github.io/videogpt/index.html}.
\end{abstract}

\section{Introduction}

Deep generative models of multiple types~\citep{kingma2013auto, goodfellow2014generative, oord2016pixel, dinh2016density} have seen incredible progress in the last few years on multiple modalities including natural images~\citep{oord2016conditional, zhang2019self, brock2018large, kingma2018glow, ho2019flow++, karras2017progressive,  karras2019style, van2017neural, razavi2019generating, vahdat2020nvae, ho2020denoising, chen2020generative, ramesh2021zero}, audio waveforms conditioned on language features~\citep{oord2016wavenet, oord2017parallel, prenger2019waveglow, binkowski2019high}, natural language in the form of text~\citep{radford2019language, brown2020language}, and music generation~\citep{dhariwal2020jukebox}. These results have been made possible thanks to fundamental advances in deep learning architectures~\citep{he2015deep, oord2016pixel, oord2016conditional, vaswani2017attention, zhang2019self, menick2018generating} as well as the availability of compute resources~\citep{jouppi2017datacenter, amodei2018ai} that are more powerful and plentiful than a few years ago.

While there have certainly been impressive efforts to model videos~\cite{vondrick2016generating, kalchbrenner2016video, tulyakov2018mocogan, clark2019adversarial}, high-fidelity natural videos is one notable modality that has not seen the same level of progress in generative modeling as compared to images, audio, and text. This is reasonable since the complexity of natural videos requires modeling correlations across both space and time with much higher input dimensions. Video modeling is therefore a natural next challenge for current deep generative models. The complexity of the problem also demands more compute resources which can also be deemed as one important reason for the {\it relatively} slow progress in generative modeling of videos.

Why is it useful to build generative models of videos? Conditional and unconditional video generation implicitly addresses the problem of video prediction and forecasting. Video prediction~\citep{srivastava2015unsupervised, finn2016unsupervised, kalchbrenner2017video, sonderby2020metnet} can be seen as learning a generative model of future frames conditioned on the past frames. Architectures developed for video generation can be useful in forecasting applications for weather prediction~\cite{sonderby2020metnet}, autonomous driving (for e.g., such as predicting the future in more semantic and dense abstractions like segmentation masks~\citep{Luc_2017_ICCV}). Finally, building generative models of the world around us is considered as one way to measure our understanding of physical common sense and predictive intelligence~\citep{lake2015human}.

Multiple classes of generative models have been shown to produce strikingly good samples such as autoregressive models~\citep{oord2016pixel, oord2016conditional, parmar2018image, menick2018generating, radford2019language, chen2020generative}, generative adversarial networks (GANs)~\citep{goodfellow2014generative, radford2015unsupervised}, variational autoencoders (VAEs)~\citep{kingma2013auto, kingma2016iaf,mittal2017sync,marwah2017attentive,vahdat2020nvae, child2020very}, Flows~\citep{dinh2014nice, dinh2016density, kingma2018glow, ho2019flow++}, vector quantized VAE (VQ-VAE)~\citep{van2017neural, razavi2019generating, ramesh2021zero}, and lately diffusion and score matching models~\citep{sohl2015deep, song2019generative, ho2020denoising}. These different generative model families have their tradeoffs across various dimensions: sampling speed, sample diversity, sample quality, optimization stability, compute requirements, ease of evaluation, and so forth. Excluding score-matching models, at a broad level, one can group these models into likelihood-based (PixelCNNs, iGPT, NVAE, VQ-VAE, Glow), and adversarial generative models (GANs). The natural question is: What is a good model class to pick for studying and scaling video generation?

\begin{figure*}[!ht]
    \centering
    \includegraphics[width=0.8\textwidth]{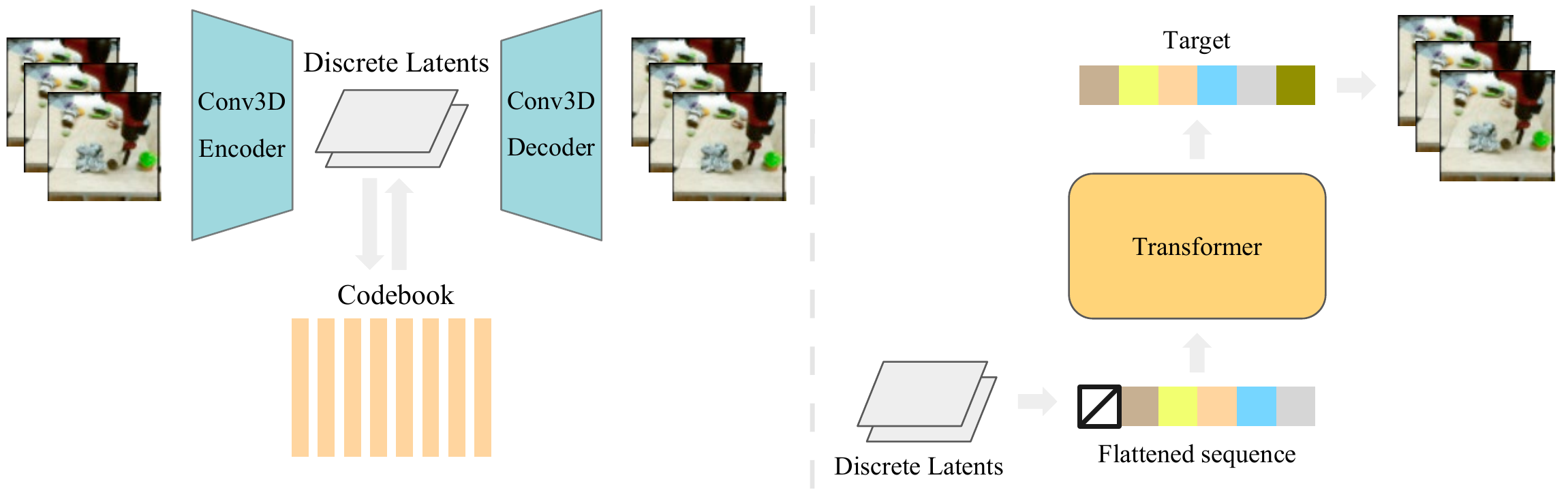}
    \caption{We break down the training pipeline into two sequential stages: training VQ-VAE (Left) and training an autoregressive transformer in the latent space (Right). The first stage is similar to the original VQ-VAE training procedure. During the second stage, VQ-VAE encodes video data to latent sequences as training data for the prior model. For inference, we first sample a latent sequence from the prior, and then use VQ-VAE to decode the latent sequence to a video sample. }
    \label{fig:videogen_architecture}
\end{figure*}

First, we make a choice between likelihood-based and adversarial models. Likelihood-based models are convenient to train since the objective is well understood, easy to optimize across a range of batch sizes, and easy to evaluate. Given that videos already present a hard modeling challenge due to the nature of the data, we believe likelihood-based models present fewer difficulties in the optimization and evaluation, hence allowing us to focus on the architecture modeling\footnote{It is not the focus of this paper to say likelihood models are better than GANs for video modeling. This is purely a design choice guided by our inclination to explore likelihood based generative models and non-empirically established beliefs with respect to stability of training.}. Next, among likelihood-based models, we pick autoregressive models simply because they have worked well on discrete data in particular, have shown greater success in terms of sample quality~\cite{ramesh2021zero}, and have well established training recipes and modeling architectures that take advantage of latest innovations in Transformer architectures~\cite{vaswani2017attention, child2019generating, ho2019axial,huang2019ccnet}.

Finally, among autoregressive models, we consider the following question: Is it better to perform autoregressive modeling in a downsampled latent space without spatio-temporal redundancies compared to modeling at the atomic level of all pixels across space and time? Below, we present our reasons for choosing the former: Natural images and videos contain a lot of spatial and temporal redundancies and hence the reason we use image compression tools such as JPEG~\citep{wallace1992jpeg} and video codecs such as MPEG~\citep{le1991mpeg} everyday. These redundancies can be removed by learning a denoised downsampled encoding of the high resolution inputs. For example, 4x downsampling across spatial and temporal dimensions results in 64x downsampled resolution so that the computation of powerful deep generative models is spent on these more fewer and useful bits. As shown in VQ-VAE~\citep{van2017neural}, even a lossy decoder can transform the latents to generate sufficiently realistic samples. This framework has in recent times produce high quality text-to-image generation models such as DALL-E~\cite{ramesh2021zero}. Furthermore, modeling in the latent space downsampled across space and time instead of the pixel space improves sampling speed and compute requirements due to reduced dimensionality.

The above line of reasoning leads us to our proposed model: VideoGPT\footnote{We note that Video Transformers~\cite{weissenborn2019scaling} also employ generative pre-training for videos using the Subscale Pixel Networks (SPN)~\cite{menick2018generating} architecture. Despite this, it is fair to use the GPT terminology for our model because our architecture more closely resembles the vanilla Transformer in a manner similar to iGPT~\cite{chen2020generative}.}, a simple video generation architecture that is a minimal adaptation of VQ-VAE and GPT architectures for videos. VideoGPT employs 3D convolutions and transposed convolutions~\citep{tran2015learning} along with axial attention~\citep{huang2019ccnet, ho2019axial} for the autoencoder in VQ-VAE, learning a downsampled set of discrete latents from raw pixels of the video frames. These latents are then modeled using a strong autoregressive prior using a GPT-like~\citep{radford2019language, child2019generating, chen2020generative} architecture. The generated latents from the autoregressive prior are then decoded to videos of the original resolution using the decoder of the VQ-VAE.

Our results are highlighted below:

1. On the widely benchmarked BAIR Robot Pushing dataset~\citep{ebert2017self}, VideoGPT can generate realistic samples that are competitive with existing methods such as TrIVD-GAN~\citep{luc2020transformation}, achieving an FVD of 103 when benchmarked with real samples, and an FVD*~\citep{razavi2019generating} of 94 when benchmarked with reconstructions.

2. In addition, VideoGPT is able to generate realistic samples from complex natural video datasets, such as UCF-101 and the Tumblr GIF dataset

3. We present careful ablation studies for the several architecture design choices in VideoGPT including the benefit of axial attention blocks, the size of the VQ-VAE latent space, number of codebooks, and the capacity (model size) of the autoregressive prior.

4. VideoGPT can easily be adapted for action conditional video generation. We present qualitative results on the BAIR Robot Pushing dataset and Vizdoom simulator~\citep{kempka2016vizdoom}. \\

\section{Background}

\subsection{VQ-VAE}
The Vector Quantized Variational Autoencoder (VQ-VAE)~\citep{van2017neural} is a model that learns to compress high dimensional data points into a discretized latent space and reconstruct them. The encoder $E(x)\rightarrow h$ first encodes $x$ into a series of latent vectors $h$ which is then discretized by performing a nearest neighbors lookup in a codebook of embeddings $C = \{e_i\}_{i=1}^K$ of size $K$. The decoder $D(e)\rightarrow \hat{x}$ then learns to reconstruct $x$ from the quantized encodings. The VQ-VAE is trained using the following objective:
\begin{equation*}
\mathcal{L} = \underbrace{\norm{x - D(e)}_2^2}_{\mathcal{L}_{\text{recon}}} + \underbrace{\norm{sg[E(x)] - e}_2^2}_{\mathcal{L}_{\text{codebook}}} + \underbrace{\beta\norm{sg[e] - E(x)}_2^2}_{\mathcal{L}_{\text{commit}}}
\end{equation*}

where $sg$ refers to a stop-gradient. The objective consists of a reconstruction loss $\mathcal{L}_{\text{recon}}$, a codebook loss $\mathcal{L}_{\text{codebook}}$, and a commitment loss $\mathcal{L}_{\text{commit}}$. The reconstruction loss encourages the VQ-VAE to learn good representations to accurately reconstruct data samples. The codebook loss brings codebook embeddings closer to their corresponding encoder outputs, and the commitment loss is weighted by a hyperparameter $\beta$ and prevents the encoder outputs from fluctuating between different code vectors. 

An alternative replacement for the codebook loss described in \cite{van2017neural} is to use an EMA update which empirically shows faster training and convergence speed. In this paper, we use the EMA update when training the VQ-VAE. 

\subsection{GPT}

GPT and Image-GPT~\citep{chen2020generative} are a class of autoregressive transformers that have shown tremendous success in modelling discrete data such as natural language and high dimensional images. These models factorize the data distribution $p(x)$ according to $p(x) = \prod_{i=1}^d p(x_i|x_{<i})$ through masked self-attention mechanisms and are optimized through maximum likelihood. The architectures employ multi-head self-attention blocks followed by pointwise MLP feedforward blocks following the standard design from \cite{vaswani2017attention}.

\begin{figure*}[h]
    \centering
    \includegraphics[width=\textwidth]{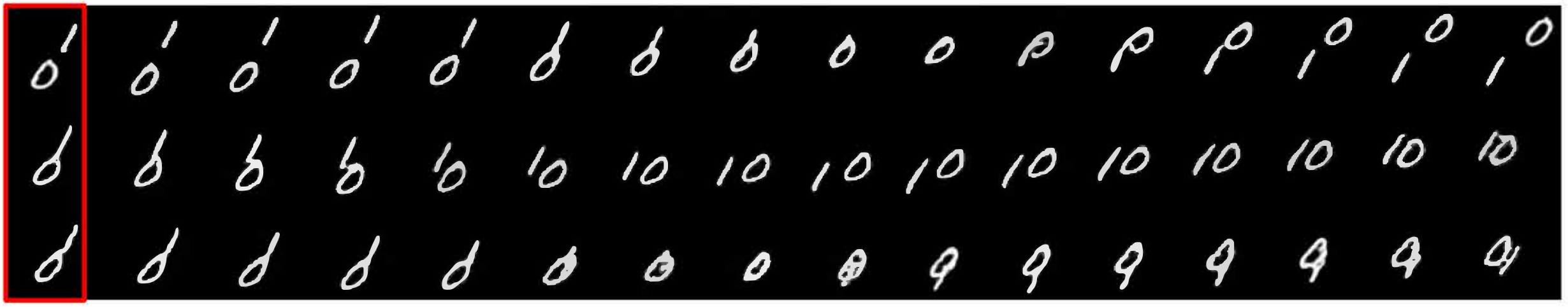}
    \caption{Moving MNIST samples conditioned on a single given frame (red). }
    \label{fig:moving_mnist}
\end{figure*}

\section{VideoGPT}
Our primary contribution is VideoGPT, a new method to model complex video data in a computationally efficient manner. An overview of our method is shown in Fig~\ref{fig:videogen_architecture}.

\begin{figure}[h]
    \centering
    \includegraphics[scale=0.3]{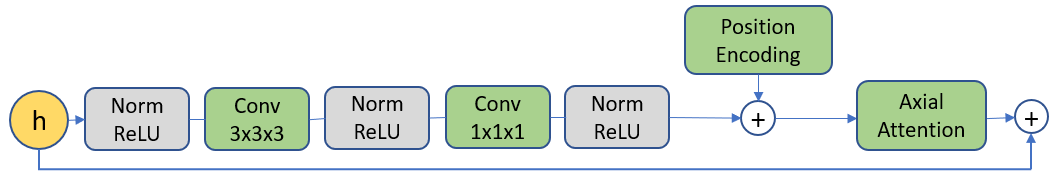}
    \caption{Architecture of the attention residual block in the VQ-VAE as a replacement for standard residual blocks.}
    \label{fig:attn_res_block}
\end{figure}

\textbf{Learning Latent Codes}
In order to learn a set of discrete latent codes, we first train a VQ-VAE on the video data. The encoder architecture consists of a series of 3D convolutions that downsample over space-time, followed by attention residual blocks. Each attention residual block is designed as shown in Fig~\ref{fig:attn_res_block}, where we use LayerNorm~\citep{ba2016layer}, and axial attention layers following \cite{huang2019ccnet,ho2019axial}.

The architecture for the decoder is the reverse of the encoder, with attention residual blocks followed by a series of 3D transposed convolution that upsample over space-time. The position encodings are learned spatio-temporal embeddings that are shared between all axial attention layers in the encoder and decoder.

\textbf{Learning a Prior}
The second stage of our method is to learn a prior over the VQ-VAE latent codes from the first stage. We follow the Image-GPT architecture for prior network, except that we add dropout layers after the feedforward and attention block layers for regularization.

Although the VQ-VAE is trained unconditionally, we can generate conditional samples by training a conditional prior. We use two types of conditioning:
\begin{itemize}
    \item \textbf{Cross Attention}: For video frame conditioning, we first feed the conditioned frames into a 3D ResNet, and then perform cross-attention on the ResNet output representation during prior network training.
    
    \item \textbf{Conditional Norms}: Similar to conditioning methods used in GANs, we parameterize the gain and bias in the transformer Layer Normalization~\citep{ba2016layer} layers as affine functions of the conditional vector. This conditioning method is used for action and class-conditioning models.
\end{itemize}

\section{Experiments}

In the following section, we evaluate our method and design experiments to answer the following questions:
\begin{itemize}
    \item Can we generate high-fidelity samples from complex video datasets?
    \item How do different architecture design choices for VQ-VAE and prior network affect performance?
\end{itemize}

\subsection{Training Details}
All image data is scaled to $[-0.5, 0.5]$ before training. For VQ-VAE training, we use random restarts for embeddings, and codebook initialization by copying encoder latents as described in \cite{dhariwal2020jukebox}. In addition, we found VQ-VAE training to be more stable (less codebook collapse) when using Normalized MSE for the reconstruction loss, where MSE loss is divided by the variance of the dataset. For all datasets except UCF-101, we train on $64\times 64$ videos of sequence length $16$. For the transformer, we train Sparse Transformers~\cite{child2019generating} with local and strided attention across space-time. Exact architecture details and hyperparameters can be found in Appendix~\ref{appendix:arch_hyperparams}. We achieve all results with a maximum of 8 Quadro RTX 6000 GPUs (24 GB memory).

\subsection{Moving MNIST}
For Moving MNIST, VQ-VAE downsamples input videos by a factor of 4 over space-time (64x total reduction), and contains two residual layers with no axial-attention. We use a codebook of $512$ codes, each $64$-dim embeddings. To learn the single-frame conditional prior, we train a conditional transformer with $384$ hidden features, $4$ heads, $8$ layers, and a ResNet-18 single frame encoder. Fig~\ref{fig:moving_mnist} shows several different generated trajectories conditioned on a single frame.

\begin{figure*}[h]
    \centering
    \includegraphics[width=\textwidth]{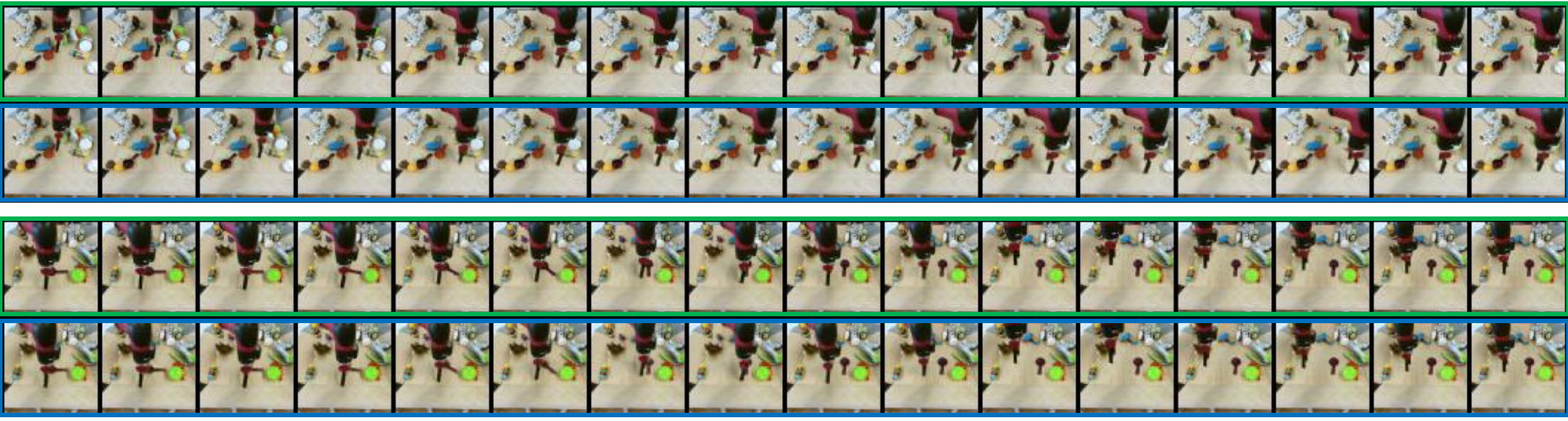}
    \caption{VQ-VAE reconstructions for BAIR Robot Pushing. The original videos are contained in green boxes and reconstructions in blue. }
    \label{fig:bair_recon}
\end{figure*}
\begin{figure*}[h]
    \centering
    \includegraphics[width=\textwidth]{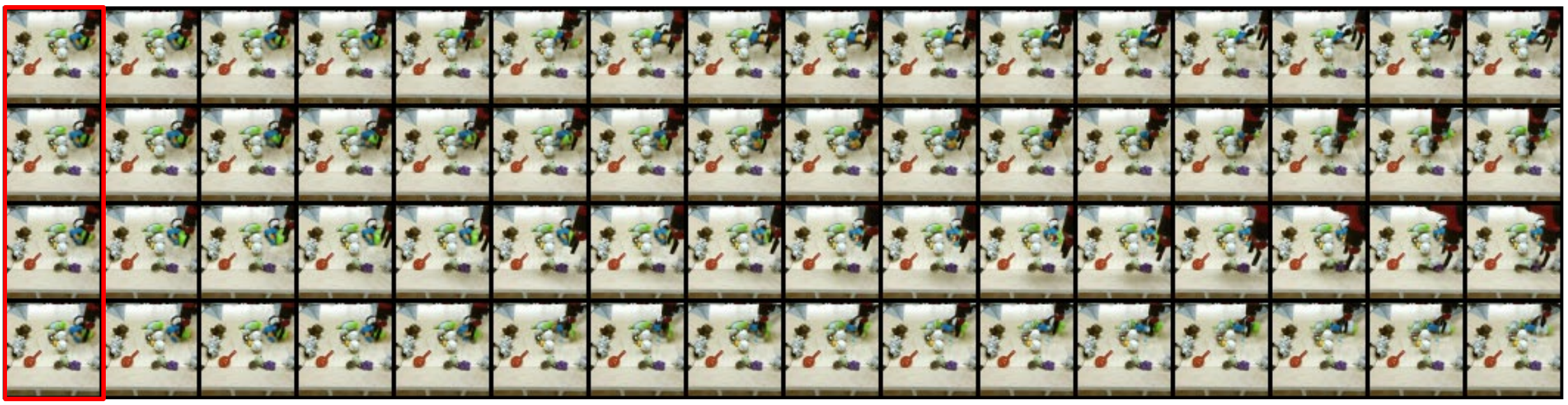}
    \caption{BAIR Robot Pushing samples from a single-frame conditioned VideoGPT model. Frames highlighting in red are conditioning frames. Although all videos follow the same starting frame, the samples eventually diverge to varied trajectories.}
    \label{fig:bair_pushing}

\end{figure*}

\begin{table}[h]
    
        \centering
    
        \caption{FVD on BAIR}
        \label{table:bair_fvd}
        \begin{tabular}{ll}
            \toprule
            Method\textsuperscript{3}                   & FVD ($\downarrow$)                \\ \midrule
            SV2P                    & $262.5$            \\
            LVT                     & $125.8$           \\
            SAVP                    & $116.4$            \\
            DVD-GAN-FP              & $109.8$            \\
            \textbf{VideoGPT (ours)}          & $103.3$            \\
            TrIVD-GAN-FP           & $103.3$             \\
            Video Transformer        & $\mathbf{94\pm 2}$ \\ \bottomrule
        \end{tabular}
\end{table}

\subsection{BAIR Robot Pushing}

For BAIR, VQ-VAE downsamples the inputs by a factor of 2x over each of height, width and time dimensions. The embedding in the latent space is a $256$-dimensional vector, which is discretized through a codebook with $1024$ codes. We use $4$ axial-attention residual blocks for the VQ-VAE encoder and a prior network with a hidden size of $512$ and $16$ layers.

Quantitatively, Table~\ref{table:bair_fvd}\footnote{SV2P~\citep{babaeizadeh2017stochastic}, SAVP~\citep{lee2018stochastic}, DVD-GAN-FP~\citep{clark2019adversarial}, Video Transformer~\citep{weissenborn2019scaling}, Latent Video Transformer (LVT)~\citep{rakhimov2020latent}, and TrIVD-GAN~\citep{luc2020transformation} are our baselines} shows FVD results on BAIR, comparing our method with prior work. Although our method does not achieve state of the art, it is able to produce very realistic samples competitive with the best performing GANs.

Qualitatively, Fig~\ref{fig:bair_recon} shows VQ-VAE reconstructions on BAIR. Fig~\ref{fig:bair_pushing} shows samples primed with a single frames. We can see that our method is able to generate realistically looking samples. In addition, we see that VideoGPT is able to sample different trajectories from the same initial frame, showing that it is not simply copying the dataset.  

\subsection{ViZDoom}
For ViZDoom, we use the same VQ-VAE and transformer architectures as for the BAIR dataset, with the exception that the transformer is trained without single-frame conditioning. We collect the training data by training a policy in each ViZDoom environment, and collecting rollouts of the final trained policies. The total dataset size consists of 1000 episodes of length 100 trajectories, split into an 8:1:1 train / validataion / test ratio. We experiment on the Health Gathering Supreme and Battle2 ViZDoom environments, training both unconditional and action-conditioned priors. VideoGPT is able to capture complex 3D camera movements and environment interactions. In addition, action-conditioned samples are visually consistent with the input action sequence and show a diverse range of backgrounds and scenarios under different random generations for the same set of actions. Samples can be found in Appendix~\ref{appendix:vizdoom_samples}.\footnote{VGAN~\citep{vondrick2016generating}, TGAN~\cite{saito2017temporal}, MoCoGAN~\citep{tulyakov2018mocogan}, Progressive VGAN~\cite{acharya2018towards}, TGAN-F~\citep{kahembwe2020lower}, TGANv2~\cite{saito2018tganv2}, DVD-GAN~\cite{clark2019adversarial} are our baselines for IS on UCF-101.}

\begin{table}[h]
            \centering
        
        \caption{IS on UCF-101}
        \label{table:ucf_is}
\begin{tabular}{@{}cc@{}}
\toprule
Method\textsuperscript{4} & IS ($\uparrow$)                      \\ \midrule
VGAN                      & $8.31 \pm 0.09$                      \\
TGAN                      & $11.85 \pm 0.07$                     \\
MoCoGAN                   & $12.42 \pm 0.03$                     \\
Progressive VGAN          & $14.56 \pm 0.05$                     \\
TGAN-F                    & \multicolumn{1}{l}{$22.91 \pm 0.19$} \\
\textbf{VideoGPT (ours)}  & \multicolumn{1}{l}{$24.69 \pm 0.30$} \\
TGANv2                    & \multicolumn{1}{l}{$28.87 \pm 0.67$} \\
DVD-GAN                   & \multicolumn{1}{l}{$\mathbf{32.97 \pm 1.7}$}  \\ \bottomrule
\end{tabular}
\end{table}

\subsection{UCF-101}
UCF-101~\citep{soomro2012ucf101} is an action classification dataset with 13,320 videos from 101 different classes. We train unconditional VideoGPT models on $16$ frame $64 \times 64$ and $128\times 128$ videos, where the original videos have their shorter side scaled to $128$ pixels, and then center cropped. 

Table~\ref{table:ucf_is} shows results comparing Inception Score\footnote{Inception Score is calculated using the code at \url{https://github.com/pfnet-research/tgan2}} (IS)~\citep{salimans2016improved} calculations against various baselines. Unconditionally generated samples are shown in Figure~\ref{fig:ucf}. Similarly observed in~\cite{clark2019adversarial}, we notice that that VideoGPT easily overfits UCF-101 with a train loss of $3.40$ and test loss of $3.12$, suggesting that UCF-101 may be too small a dataset of the relative complexity of the data itself, and more exploration would be needed on larger datasets.

\subsection{Tumblr GIF (TGIF)}
\begin{figure*}[ht]
\begin{minipage}{\textwidth}
        \centering
    \includegraphics[width=\textwidth]{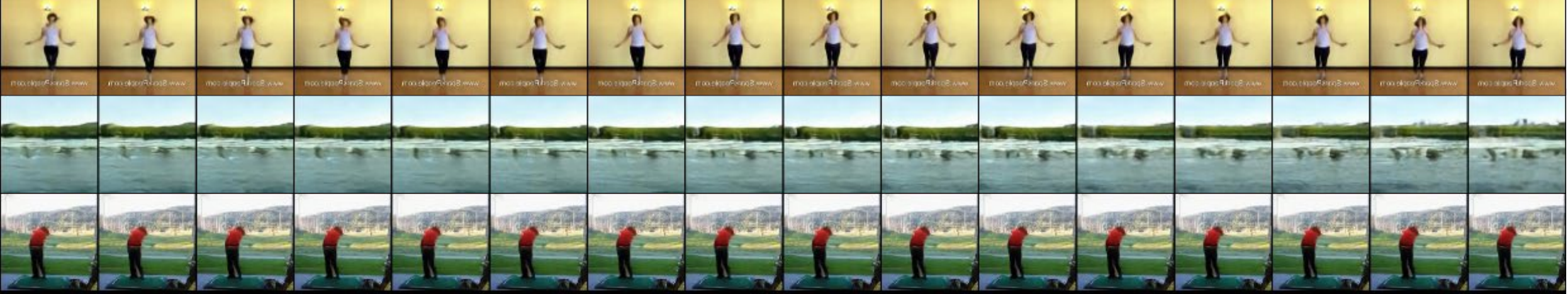}
    \caption{$128\times 128$ UCF-101 unconditional samples}
    \label{fig:ucf}
\end{minipage}
\begin{minipage}{\textwidth}
        \centering
    \includegraphics[width=\textwidth]{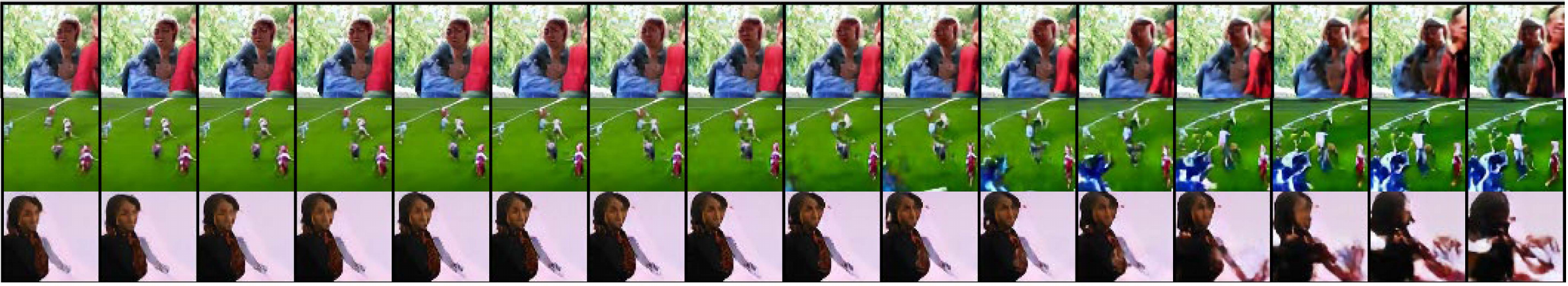}
    \caption{$64 \times 64$ TGIF unconditional samples}
    \label{fig:tgif}
\end{minipage}

\end{figure*}

TGIF~\citep{li2016tgif} is a dataset of 103,068 selected GIFs from Tumblr, totalling around 100,000 hours of video. Figure~\ref{fig:tgif} shows samples from a trained unconditional VideoGPT model. We see that the video sample generations are able to capture complex interactions, such as camera movement, scene changes, and human and object dynamics. Unlike UCF-101, VideoGPT did not overfit on TGIF with a train loss of $2.87$ and test loss $2.86$.

\subsection{Ablations}
In this section, we perform ablations on various architectural design choices for VideoGPT.

\textbf{Axial-attention in VQ-VAE increases reconstruction and generation quality.}
\begin{table}[H]
    \begin{minipage}{.5\textwidth}
        \centering

        \caption{Ablation on attention in VQ-VAE. R-FVD is with reconstructed examples}
        \label{table:abl_vqvae_attn}
        \begin{tabular}{@{}lll@{}}
            \toprule
            VQ-VAE Architecture & NMSE ($\downarrow$)       & R-FVD ($\downarrow$)   \\ \midrule
            No Attention        & 0.0041      & 15.3  \\
            With Attention      & $\mathbf{0.0033}$ & $\mathbf{14.9}$ \\ \bottomrule
        \end{tabular}
    \end{minipage}
\end{table}
We compare VQ-VAE with and without axial attention blocks as shown in Table~\ref{table:abl_vqvae_attn}. Empirically, incorporating axial attention into the VQ-VAE architecture improves reconstruction (NMSE) performance, and has better reconstruction FVD. Note that in order to take into account the added parameter count from attention layers, we increase the number of convolutional residual blocks in the "No Attention" version for better comparability. Fig~\ref{fig:bair_recon} shows samples of videos reconstructed by VQ-VAE with axial attention module. 

\textbf{Larger prior network capacity increases performance.}
\begin{table}[H]
        \centering
        \caption{Ablations comparing the number of transformer layers}
\label{table:abl_tfm_layers}
\begin{tabular}{@{}cccc@{}}
\toprule
Transformer Layers & bits/dim & FVD ($\downarrow$)  \\ \midrule
2                  & $2.84$   & $120.4 \pm 6.0$      \\
4                  & $2.52$   & $110.0 \pm 2.4$       \\
8                  & $2.39$   & $103.3 \pm 2.2$       \\
16                 & $2.05$   & $103.6 \pm 2.0$       \\ \bottomrule
\end{tabular}
\end{table}
Computational efficiency is a primary advantage of our method, where we can first use the VQ-VAE to downsample by space-time before learning an autoregressive prior. Lower resolution latents allow us to train a larger and more expressive prior network to learn complex data distributions under memory constraints. We run an ablation on the prior network size which shows that a larger transformer network produces better results. Table~\ref{table:abl_tfm_layers} shows the results of training transformers of varied number of layers on BAIR. We can see that for BAIR, our method benefits from training larger models, where the bits per dim shows substantial improvement in increasing layers, and FVD and sample quality show increments in performance up until around 8 layers. 

\textbf{A balanced temporal-spatial downsampling in VQ-VAE latent space increase performance.}
\begin{table}[H]
\begin{minipage}{.5\textwidth}
\center
\caption{Ablations comparing different VideoGPT latent sizes on BAIR. R-FVD is the FVD of VQ-VAE reconstructions, and FVD* is the FVD between samples generated by VideoGPT and samples encoded-decoded from VQ-VAE. For each partition below, the total number of latents is the same with varying amounts of spatio-temporal downsampling}
\label{table:abl_latent_size}
\begin{tabular}{@{}cccc@{}}
\toprule
Latent Size                      & R-FVD ($\downarrow$) & FVD ($\downarrow$)       & FVD* ($\downarrow$)     \\ \midrule
$4 \times 16 \times 16$          & $82.1$               & $135.4 \pm 3.7$          & $81.8 \pm 2.3$          \\
$16\times 8 \times 8$            & $108.1$              & $166.9 \pm 3.1$          & $81.6 \pm 2.2$          \\ \midrule
$8\times 16\times 16$            & $49.9$               & $124.7 \pm 2.7$          & $90.2 \pm 2.4$          \\ \midrule
$1 \times 64 \times 64$          & $41.6$               & $126.7 \pm 3.1$          & $98.1 \pm 3.6$          \\
$4 \times 32 \times 32$          & $28.3$               & $104.6 \pm 2.7$          & $90.6 \pm 2.7$          \\
$16 \times 16 \times 16$         & $32.8$               & $113.4 \pm 2.5$          & $94.9 \pm 1.7$          \\ \midrule
$2 \times 64 \times 64$          & $22.4$               & $124.3 \pm 1.4$          & $104.4 \pm 2.5$         \\
$\mathbf{8 \times 32 \times 32}$ & $\mathbf{14.9}$      & $\mathbf{103.6 \pm 2.0}$ & $\mathbf{94.6 \pm 1.5}$ \\ \midrule
$4 \times 64 \times 64$          & $15.7$               & $109.4 \pm 2.1$          & $102.3 \pm 2.8$         \\
$16 \times 32 \times 32$         & $10.1$               & $118.4 \pm 3.2$          & $113.8 \pm 3.3$         \\ \bottomrule
\end{tabular}
\end{minipage}

\end{table}

A larger downsampling ratio results in a smaller latent code size, which allows us to train larger and more expressive prior models. However, limiting the expressivity of the discrete latent codes may introduce a bottleneck that results in poor VQ-VAE reconstruction and sample quality. Thus, VideoGPT presents an inherent trade-off between the size of the latents, and the allowed capacity of prior network. Table~\ref{table:abl_latent_size} shows FVD results from training VideoGPT on varying latent sizes for BAIR. We can see that larger latents sizes have better reconstruction quality (lower R-FVD), however, the largest latents $16 \times 32 \times 32$ does not perform the best sample-quality-wise due to limited compute constraints on prior model size. On the other hand, the smallest set of latents $4\times 16 \times 16$ and $16 \times 8 \times 8$ have poor reconstruction quality and poor samples. There is a sweet-spot in the trade-off at around $8\times 32 \times 32$ where we observe the best sample quality.

In addition to looking at the total number of latents, we also investigate the appropriate downsampling for each latent resolution. Each partition in Table~\ref{table:abl_latent_size} shows latent sizes with the same number of total latents, each with different spatio-temporal downsampling allocations. Unsurprisingly, we find that a balance of downsampling ratio ($2\times 2 \times 2$, corresponding to latent size $8\times 32 \times 32$) between space and time is the best, as opposed to downsampling over only space or only time. 

\textbf{Further increasing the number of latent codes does not affect performance.}
\begin{table}[H]
\begin{minipage}{.5\textwidth}
\centering
\caption{Ablations comparing the number of codebook codes}
\label{table:abl_n_codes}
\begin{tabular}{@{}cccc@{}}
\toprule
\# of Codes & R-FVD ($\downarrow$) & FVD ($\downarrow$) & bits/dim \\ \midrule
256         & $18.2$               & $103.8 \pm 3.7$    & $1.55$      \\
1024        & $14.9$               & $103.6 \pm 2.0$    & $2.05$      \\
4096        & $11.3$               & $103.9 \pm 5.1$    & $2.60$      \\ \bottomrule
\end{tabular}
\end{minipage}
\end{table}
In Table~\ref{table:abl_n_codes}, we show experimental results for running VideoGPT with different number of codes in the codebooks. For all three runs, the VQ-VAE latent code vector has size $8 \times 32 \times 32$. In the case of BAIR, we find that reconstruction quality improves with increasing the number of codes due to better expressivity in the discrete bottleneck. However, they ultimately do not affect sample quality. This may be due to the fact that in the case of BAIR, using 256 codes surpasses a base threshold for generation quality.

\textbf{Using one VQ-VAE codebook instead of multiple improves performance.}
\begin{table}[H]
\caption{Ablations comparing the number of codebooks}
\label{table:abl_n_codebooks}
\begin{tabular}{@{}cccc@{}}
\toprule
Latent Size                    & R-FVD ($\downarrow$) & FVD ($\downarrow$) & bits/dim \\ \midrule
$\mathbf{8\times32\times32\times 1}$    & $\mathbf{14.9}$               & $\mathbf{103.6 \pm 2.0}$    & $\mathbf{2.05}$      \\
$16\times 16\times 16\times 2$ & $17.2$               & $106.3 \pm 1.7$                   & $2.41$                \\
$8\times 16\times 16 \times 4$ & $17.7$               & $131.4 \pm 2.9$    & $2.68$     \\
$4\times 16\times 16 \times 8$ & $23.1$               & $135.7 \pm 3.3$    & $2.97$     \\ \bottomrule
\end{tabular}
\end{table}
In our main results, we use one codebook for VQ-VAE. In Table~\ref{table:abl_n_codebooks}, we compare VideoGPT with different number of codebooks. Specifically, multiple codebooks is implemented by multiplying VQ-VAE's encode output channel dimension by $C$ times, where $C$ is the number of codebooks. The encoder output is then sliced along channel dimension, and each slice is quantized through a separate codebook. As a result, the size of the discrete latents are of dimension $T \times H \times W \times C$, as opposed to $T\times H\times W$ when using a single codebook. Generally, multiple codebooks may be more favorable over increasing the downsampling resolution as multiple codebooks allows a combinatorially better scaling in bottleneck complexity. In our experiments, we increase the number of codebooks, and reduce spatio-temporal resolutions on latent sizes to keep the size of the latent space constant. We see that increasing the number of codebooks worsens sample quality performance, and the best results are attained at the highest resolution with one codebook. Nevertheless, incorporating multiple codebooks might shows its advantage when trained with a larger dataset or a different VQ-VAE architecture design. 

\section{Related Work}

% High level points:
{\bf Video Prediction}
The problem of video prediction~\citep{srivastava2015unsupervised} is quite related to video generation in that the latter is one way to solve the former.  
Plenty of methods have been proposed for video prediction on the BAIR Robot dataset~\citep{finn2016unsupervised, ebert2017self, babaeizadeh2017stochastic, denton2017unsupervised, denton2018stochastic, lee2018stochastic} where the future frames are predicted given the past frame(s) and (or) action(s) of a robot arm moving across multiple objects thereby benchmarking the ability of video models to capture object-robot interaction, object permanance, robot arm motion, etc. Translating videos to videos is another paradigm to think about video prediction with a prominent example being \texttt{vid2vid}~\cite{wang2018video}. The \texttt{vid2vid} framework uses automatically generated supervision from more abstract information such as semantic segmentation~\citep{Luc_2017_ICCV} masks, keypoints, poses, edge detectors, etc to further condition the GAN based video translation setup.

{\bf Video Generation}
Most modern generative modeling architectures allow for easy adaptation of unconditional video generation to conditional versions through conditional batch-norm~\citep{brock2018large}, concatenation~\citep{salimans2017pixelcnn++, oord2016conditional}, etc. Video Pixel Networks~\citep{kalchbrenner2017video} propose a convolutional LSTM based encoding of the past frames to be able to generate the next frame pixel by pixel autoregressively with a PixelCNN~\citep{oord2016conditional} decoder. The architecture serves both as a video generative as well as predictive model, optimized through log-likelihood loss at the pixel level. Subscale Video Transformers~\citep{weissenborn2019scaling} extend the idea of Subscale Pixel Networks~\citep{menick2018generating} for video generation at the pixel level using the subscale ordering across space and time. However, the sampling time and compute requirements are large for these models. In the past, video specific architectures have been proposed for GAN based video generation with primitive results by \cite{vondrick2016generating}. Recently, DVD-GAN proposed by \cite{clark2019adversarial} adopts a BigGAN like architecture for videos with disentangled (axial) non-local~\citep{wang2017non} blocks across space and time. They present a wide range of results, unconditional, past frame(s) conditional, and class conditional video generation. 

Other examples of prior work with video generation of GANs include \cite{saito2017temporal}, \cite{tulyakov2018mocogan}, \cite{acharya2018towards}, \cite{yushchenko2019markov}. In addition, \cite{saito2018tganv2} and \cite{kahembwe2020lower} propose more scalable and efficient GAN models for training on less compute. Our approach builds on top of VQ-VAE~\citep{van2017neural} by adapting it for video generation. A clean architecture with VQ-VAE for video generation has not been presented yet and we hope VideoGPT is useful from that standpoint. While VQ-VAE-2~\citep{razavi2019generating} proposes using multi-scale hierarchical latents and SNAIL blocks~\cite{chen2017pixelsnail} (and this setup has been applied to videos in \cite{walker2021predicting}), the pipeline is inherently complicated and hard to reproduce. For simplicity, ease of reproduction and presenting the first VQ-VAE based video generation model with minimal complexity, we stick with a single scale of discrete latents and transformers for the autoregressive priors, a design choice also adopted in DALL-E~\cite{ramesh2021zero}.

\section{Conclusion}
We have presented VideoGPT, a new video generation architecture adapting VQ-VAE and Transformer models typically used for image generation to the domain of videos with minimal modifications. We have shown that VideoGPT is able to synthesize videos that are competitive with state-of-the-art GAN based video generation models. We have also presented ablations on key design choices used in VideoGPT which we hope is useful for future design of architectures in video generation.

\section*{Acknowledgement}
The work was in part supported by NSF NRI Grant \#2024675 and by Berkeley Deep Drive.

\bibliography{main}
\bibliographystyle{icml2021}

\clearpage
\appendix
\onecolumn
\section{Architecture Details and Hyperparameters}
\label{appendix:arch_hyperparams}
\subsection{VQ-VAE Encoder and Decoder}

\begin{table}[h]
\caption{Hyperparameters of VQ-VAE encoder and decoder models for each dataset}
\begin{tabular}{@{}llcc@{}}
\toprule
                                                           & Moving MNIST           & \multicolumn{1}{l}{BAIR / RoboNet / ViZDoom} & \multicolumn{1}{l}{UCF-101 / TGIF} \\ \midrule
\multicolumn{1}{l|}{Input size}                            & $16\times 64\times 64$ & $16\times 64\times 64$                       & $16\times 64\times 64$             \\
\multicolumn{1}{l|}{Latent size}                           & $4\times16\times16$    & $8\times32\times 32$                         & $4\times32\times32$                \\
\multicolumn{1}{l|}{$\beta$ (commitment loss coefficient)} & 0.25                   & 0.25                                         & 0.25                               \\
\multicolumn{1}{l|}{Batch size}                            & 32                     & 32                                           & 32                                 \\
\multicolumn{1}{l|}{Learning rate}                         & $7\times 10^{-4}$      & $7\times 10^{-4}$                            & $7\times 10^{-4}$                  \\
\multicolumn{1}{l|}{Hidden units}                          & 240                    & 240                                          & 240                                \\
\multicolumn{1}{l|}{Residual units}                        & 128                    & 128                                          & 128                                \\
\multicolumn{1}{l|}{Residual layers}                       & 2                      & 4                                            & 4                                  \\
\multicolumn{1}{l|}{Uses attention}                        & No                     & Yes                                          & Yes                                \\
\multicolumn{1}{l|}{Codebook size}                         & 512                    & 1024                                         & 1024                               \\
\multicolumn{1}{l|}{Codebook dimension}                    & 64                     & 256                                          & 256                                \\
\multicolumn{1}{l|}{Encoder filter size}                   & 3                      & 3                                            & 3                                  \\
\multicolumn{1}{l|}{Upsampling conv filter size}           & 4                      & 4                                            & 4                                  \\
\multicolumn{1}{l|}{Training steps}                                             & 20k                    & 100K                                         & 100K                               \\ \bottomrule
\end{tabular}
\end{table}

\subsection{Prior Networks}

\begin{table}[h]
\caption{Hyperparameters of prior networks for each dataset}
\begin{tabular}{@{}lllll@{}}
\toprule
                                             & Moving MNIST            & BAIR / RoboNet              & ViZDoom                     & UCF-101 / TGIF          \\ \midrule
\multicolumn{1}{l|}{Input size}              & $4 \times 16 \times 16$ & $8 \times 32 \times 32$     & $8\times 32 \times 32$     & $4 \times 32 \times 32$ \\
\multicolumn{1}{l|}{Conditional sizes}       & $1\times 64\times64$    & $3 \times 64 \times 64$, 64 & $60$ (HGS), $315$ (Battle2) & n/a                     \\
\multicolumn{1}{l|}{Batch size}              & 32                      & 32                          & 32                          & 32                      \\
\multicolumn{1}{l|}{Learning rate}           & $3\times 10^{-4}$       & $3\times 10^{-4}$           & $3\times 10^{-4}$           & $3\times 10^{-4}$       \\
\multicolumn{1}{l|}{Vocabulary size}         & 512                     & 1024                        & 1024                        & 1024                    \\
\multicolumn{1}{l|}{Attention heads}         & 4                       & 4                           & 4                           & 8                       \\
\multicolumn{1}{l|}{Attention layers}        & 8                       & 16                          & 16                          & 20                      \\
\multicolumn{1}{l|}{Embedding size}          & 192                     & 512                         & 512                         & 1024                    \\
\multicolumn{1}{l|}{Feedforward hidden size} & 384                     & 2048                        & 2048                        & 4096                    \\
\multicolumn{1}{l|}{Resnet depth}            & 18                      & 34                          & n/a                         & n/a                     \\
\multicolumn{1}{l|}{Resnet units}            & 512                     & 512                         & n/a                         & n/a                     \\
\multicolumn{1}{l|}{Dropout}                 & 0.1                     & 0.2                         & 0.2                         & 0.2                     \\
\multicolumn{1}{l|}{Training steps}          & 80k                     & 150K                        & 150K                        & 200K / 600K             \\ \bottomrule
\end{tabular}
\end{table}
\clearpage
\section{ViZDoom Samples}
\label{appendix:vizdoom_samples}
\begin{figure}[H]
    \centering
    \includegraphics[width=.9\textwidth]{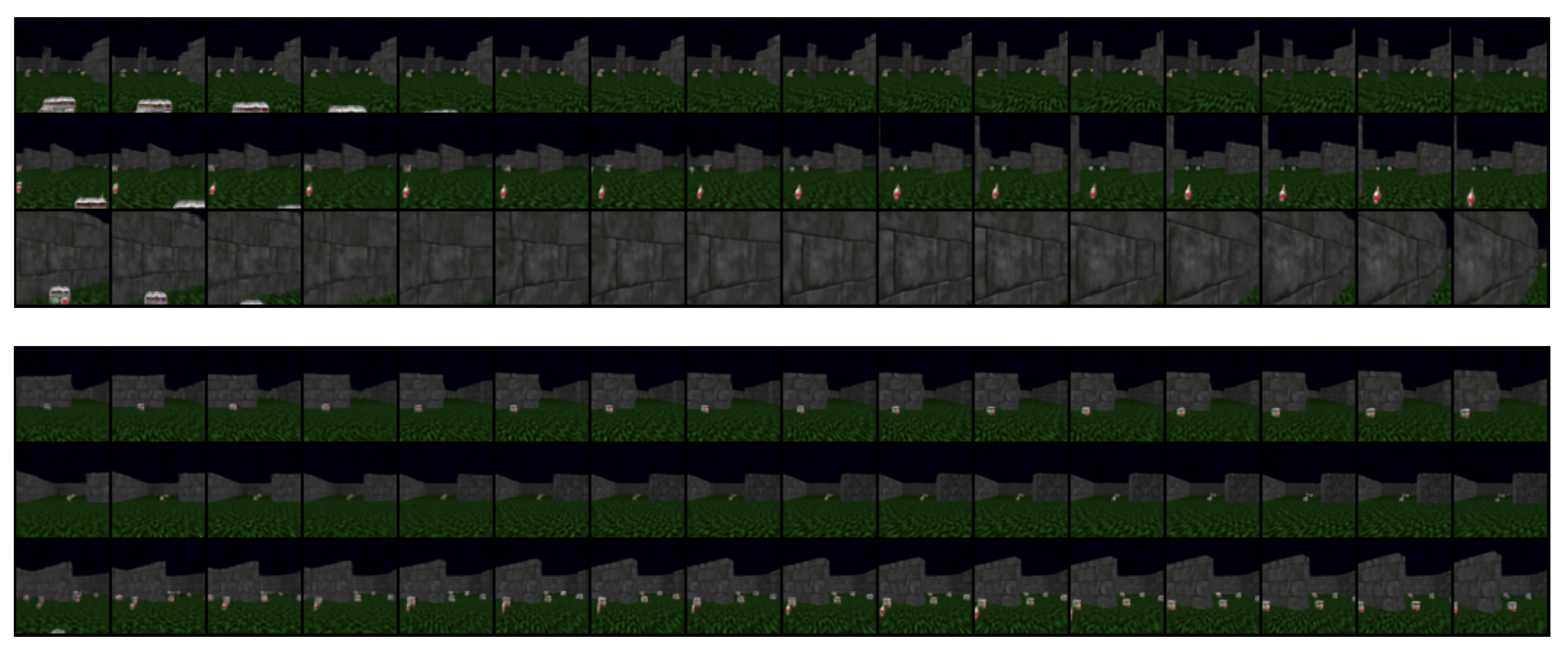}
    \caption{Samples for ViZDoom health gathering supreme environment. (Top) shows unconditionally generated samples. (Bottom) shows samples conditioned on the same action sequence (turn right and go straight).}
    \label{fig:hgs}
\end{figure}

\begin{figure}[H]
    \centering
    \includegraphics[width=.9\textwidth]{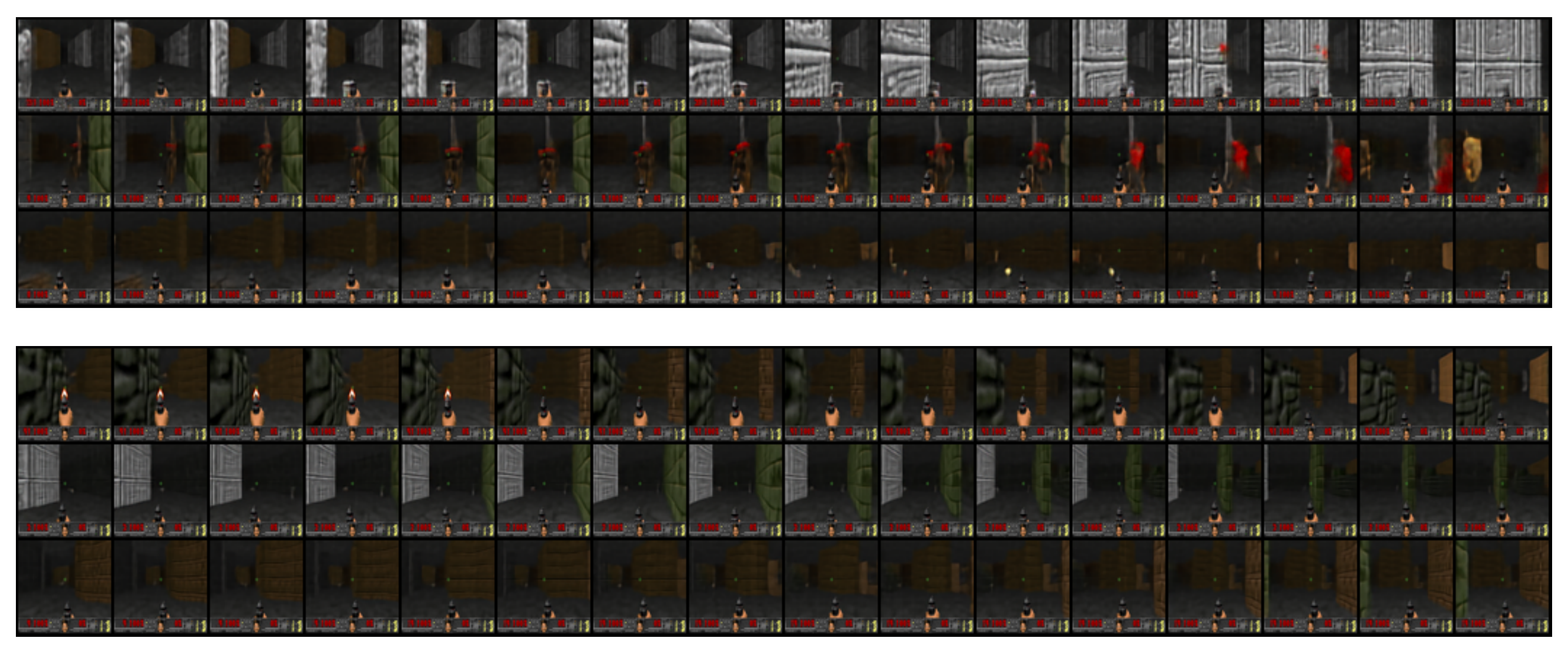}
    \caption{Samples for ViZDoom battle2 environment. (Top) shows unconditionally generated samples. (Bottom) shows three samples conditioned on the same action sequence (moving forward and right).}
    \label{fig:battle2}
\end{figure}

\end{document}